\documentclass[cameraready]{Interspeech}

\title{Enhancing Audio Reasoning via Semantic Summary Prediction \thanks{Accepted at Interspeech 2026}}

\author[affiliation={1, 2}, correspondingauthor]{Francesco}{Bonzi}
\author[affiliation={1, 2}]{Pooneh}{Mousavi}
\author[affiliation={3, 2}]{Cem}{Subakan}
\author[affiliation={1, 2}]{Mirco}{Ravanelli}

\address{
    $^1$ Concordia University, Canada \\
    $^2$ Mila - Quebec AI Institute, Canada \\
    $^3$ Université Laval, Canada
}

\email{francesco.bonzi@mila.quebec}


\usepackage{comment}
\usepackage{cite}

\begin{document}

\maketitle

\begin{abstract}
Large Audio Language Models (LALMs) perform well on complex question answering but often show a reasoning gap, where explicit Chain-of-Thought (CoT) reduces accuracy compared to direct answers. We hypothesize that long reasoning sequences shift attention away from the audio input. To address this, we propose SPARE (Semantic Prediction for Audio REasoning), which introduces a register token aligned with the final conclusion using a cosine similarity loss with a Sentence-BERT embedding. This conditions the model’s latent space with the target semantic goal before reasoning begins. Experiments on MMAU and MMAR with SALMONN show improved zero-shot reasoning and stronger early attention to audio without additional inference cost.\footnote{The implementation of SPARE is open-source and can be accessed at https://github.com/FrancescoBonzi/SPARE}
\end{abstract}

\section{Introduction}

\begin{figure*}[t]
    \vspace{-110pt}
    \centering
    \includegraphics[width=0.9\textwidth]{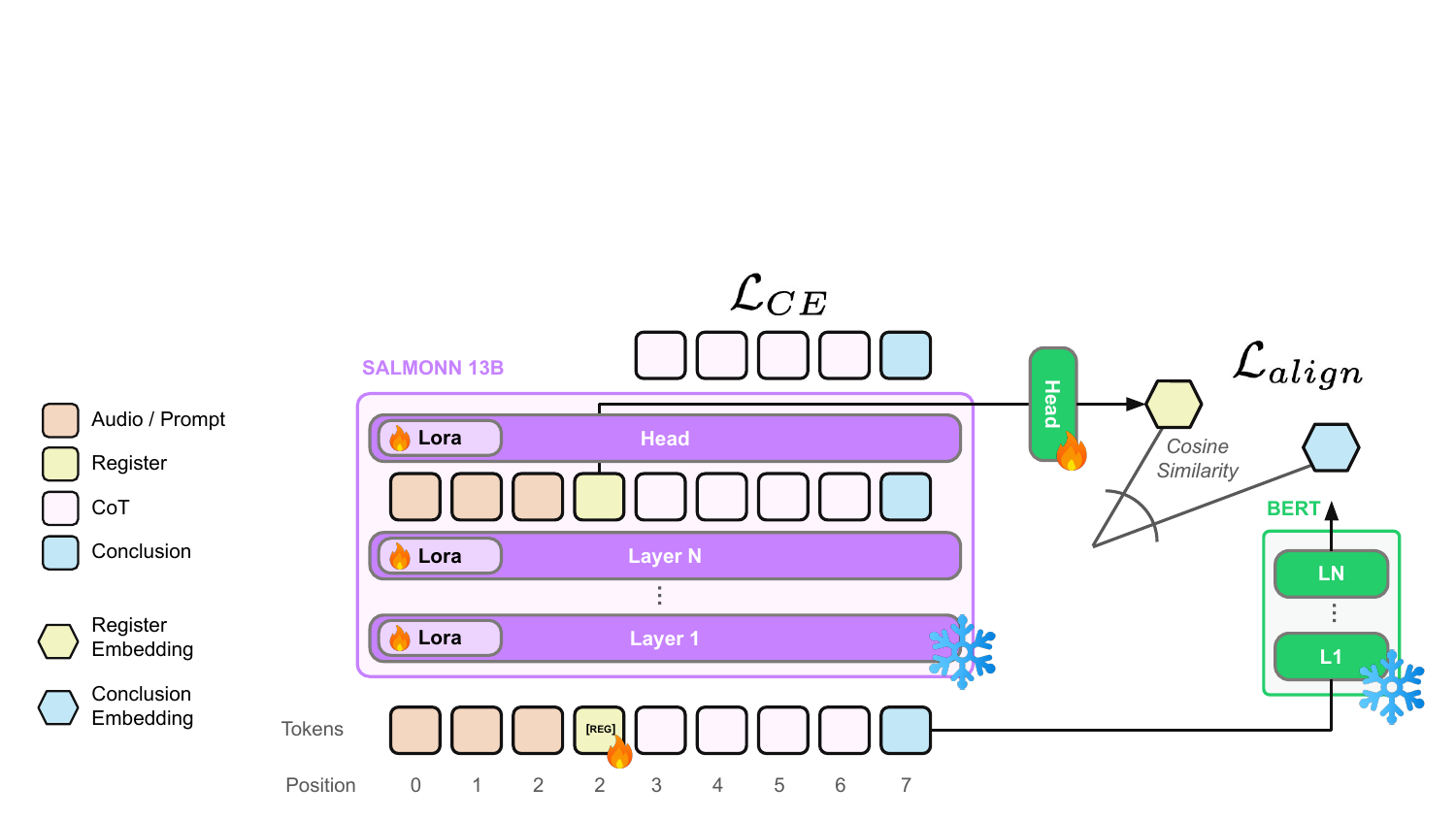}
    
    \caption{Overview of the SPARE training framework. The input sequence consists of audio/prompt tokens followed by a register token $[REG]$, the Chain-of-Thought (CoT) reasoning tokens, and the final conclusion. The training pipeline illustrates how the SALMONN-based LALM is regularized during fine-tuning. A cosine similarity alignment loss is enforced between the latent state $h_{[REG]}$ of the register token and the Sentence-BERT embedding $v_C$ of the conclusion.This encourages the model to encode the semantic goal early in the reasoning process}
    \label{fig:methodology_overview}
\end{figure*}

The integration of pre-trained audio encoders with Large Language Models (LLMs) has led to the development of powerful Large Audio Language Models (LALMs) capable of answering complex environmental and speech understanding tasks \cite{tang_salmonn_2024, gogoel_audio_2025, xu_qwen25-omni_2025, comanici2025gemini, li2025baichuan , wu_audio-thinker_2025,wu2025step,xu2025qwen3,deshmukh2025mellow}. While Chain-of-Thought (CoT) prompting has been widely adopted under the assumption that structured reasoning improves performance in the text-only domain \cite{zhouleast, yao2023tree,wei2022chain,kojima2022large}, its extension to the audio modality remains a significant challenge \cite{xie_audio-reasoner_2025, li2025reinforcement, chen2025audio,ma2025audio}.

Current LALMs often exhibit a reasoning gap: as the model generates long intermediate reasoning steps, its internal focus frequently shifts away from the acoustic input toward the linguistic tokens it has already produced. As a result, the model may ignore the audio signal and rely excessively on the generated text, producing fluent but poorly grounded answers. This often leads to lower accuracy compared to direct-answer prompting \cite{ma2025audio, yang2025sakura}. While prior efforts have attempted to address this issue, they typically rely on black-box scaling with massive supervised datasets \cite{xie_audio-reasoner_2025} or depend heavily on computationally intensive reinforcement learning frameworks \cite{wu_audio-thinker_2025}. To the best of our knowledge, there are currently no approaches that focus on architectural or training-time regularization strategies to improve reasoning in LALMs.

Recent work in natural language processing (NLP) has explored extensions to the standard next-token prediction objective to better capture long-range dependencies and planning during generation. A particularly promising direction is MuToR (Multi-Token Prediction with Registers) \cite{gerontopoulos2025}, which augments the training sequence with learnable register tokens. Each register is trained to predict future targets at different offsets, encouraging the model to develop internal “planning” representations while remaining compatible with off-the-shelf pretrained models. Another related approach \cite{mahajan2025multitokenpredictionpretrainingllms} introduces an auxiliary prediction head that learns to predict a compact representation of long-term future tokens. Inspired by these developments, we investigate how multi-token prediction strategies can be adapted to improve reasoning and temporal grounding in audio–language models. In particular, we draw inspiration from both register-based and summary-based frameworks, extending the idea beyond local next-token prediction toward learning representations that capture global latent conclusions.

We propose SPARE, a fine-tuning strategy that encourages stronger audio integration during explicit reasoning. We use SALMONN 13B \cite{tang_salmonn_2024} as our base architecture, a widely used open-source audio–language model. While newer models such as Audio Flamingo 3 \cite{gogoel_audio_2025} and Qwen-2.5-Omni \cite{xu_qwen25-omni_2025} achieve strong performance, they are trained with large amounts of instruction and reasoning data. In contrast, SALMONN has not been exposed to specialized reasoning supervision in our training set, making it a suitable platform for evaluating the effect of our reasoning-focused fine-tuning approach.

Our approach injects a specialized summary register token at the start of the CoT sequence. During training, we enforce a cosine similarity loss between the register’s final hidden state and a Sentence-BERT-derived semantic embedding \cite{wang2020minilmdeepselfattentiondistillation} of the final conclusion. By forcing the register token to align with a terminal semantic goal, rather than simply predicting the next immediate word or a local offset, as in MuToR, it acts as a semantic sieve. This mechanism encourages the self-attention layers to ignore irrelevant background information and selectively extract the acoustic features necessary to support the final conclusion. Consequently, the register token provides a top-down guidance signal that anchors the model’s internal state before the first token of the reasoning chain is generated. This explains our observation that the model becomes more selectively attentive to the audio signal in the early layers, providing a grounded foundation that prevents the logical drift or hallucinations typically observed during extended Chain-of-Thought reasoning. Additionally, SPARE requires no changes to the model architecture or inference procedure. The auxiliary parameters used for alignment are discarded after training, ensuring that improvements in zero-shot reasoning come with zero additional computational overhead. Through an analysis of self-attention maps, we demonstrate that this alignment encourages the model to attend more strongly to the audio signal in early layers, effectively regularizing the training process and improving both interpretability and audio grounding.

Our contributions are as follows:

\begin{itemize}
\item We introduce the SPARE framework, a fine-tuning regularization strategy that aligns early-stage latent representations with the final semantic goal, requiring no additional computational overhead at inference.
\item We demonstrate that SPARE significantly improves zero-shot reasoning performance on the MMAU and MMAR benchmarks.
\item Through an analysis of self-attention maps, we provide mechanistic evidence that our alignment loss encourages stronger attention to audio features in early layers, helping the model remain grounded in the acoustic input and reducing hallucinations.
\end{itemize}

\section{Methodology}

\subsection{Base Architecture}

We utilize the \textbf{SALMONN} framework \cite{tang_salmonn_2024} as our base Large Audio Language Model (LALM). SALMONN employs a dual-encoder strategy, utilizing Whisper \cite{radford_robust_2022} for speech and BEATs \cite{chen_beats_2022-1} for non-speech audio features. These are integrated via a Q-Former and a linear projection layer into a pre-trained Vicuna Large Language Model (LLM) backbone. We select SALMONN 13B as it represents a robust, open-source framework that has not been previously exposed to the reasoning-rich AF-Think dataset, allowing for a clean evaluation of our proposed fine-tuning strategy.

\subsection{Structured Reasoning and Register Injection}

To facilitate complex reasoning, we leverage AF-Think \cite{gogoel_audio_2025} labels applied to a subset of YouTube8M~\cite{gogoel_audio_2025}. Our dataset comprises 160k training and 40k validation samples. Following the methodology of Audio Reasoner \cite{xie_audio-reasoner_2025}, ground-truth targets are structured into four sequential chapters: summary, caption, reasoning, and conclusion. This multi-stage format serves as a concise CoT, which has been shown to significantly enhance reasoning performance~\cite{gogoel_audio_2025}.

We propose injecting a specialized \textbf{register token} $[REG]$ (introduced by \cite{darcet2024visiontransformersneedregisters} and exploited by \cite{gerontopoulos2025}) to serve as a latent semantic bottleneck. While register tokens have been investigated as structural placeholders in vision and language domains, they have not yet been explored within the unique dynamics of multi-modal audio reasoning. Prior configurations leverage registers primarily for local, short-range prediction offsets \cite{gerontopoulos2025}; in contrast, we explicitly repurpose the register token as a sequence-initial ``semantic bridge'' designed to combat attention drift away from long acoustic sequences. Unlike standard next-token prediction, we place this token immediately after the input context and before the reasoning chain begins. The full input sequence $\mathbf{X}$ is formatted as follows:
$$\mathbf{X} = [\text{Question}, \text{Audio}, \text{Choices}, [REG], \text{CoT}, \text{Conclusion}]$$

where \textbf{CoT} encompasses the summary, caption, and reasoning blocks. During training, the model must predict the entire text sequence, but the $[REG]$ token acts as a ``look-ahead" anchor that is regularized to capture the semantic essence of the final answer before the reasoning steps are generated.

\begin{figure}[t] 
    \centering
    \includegraphics[width=0.45\textwidth]{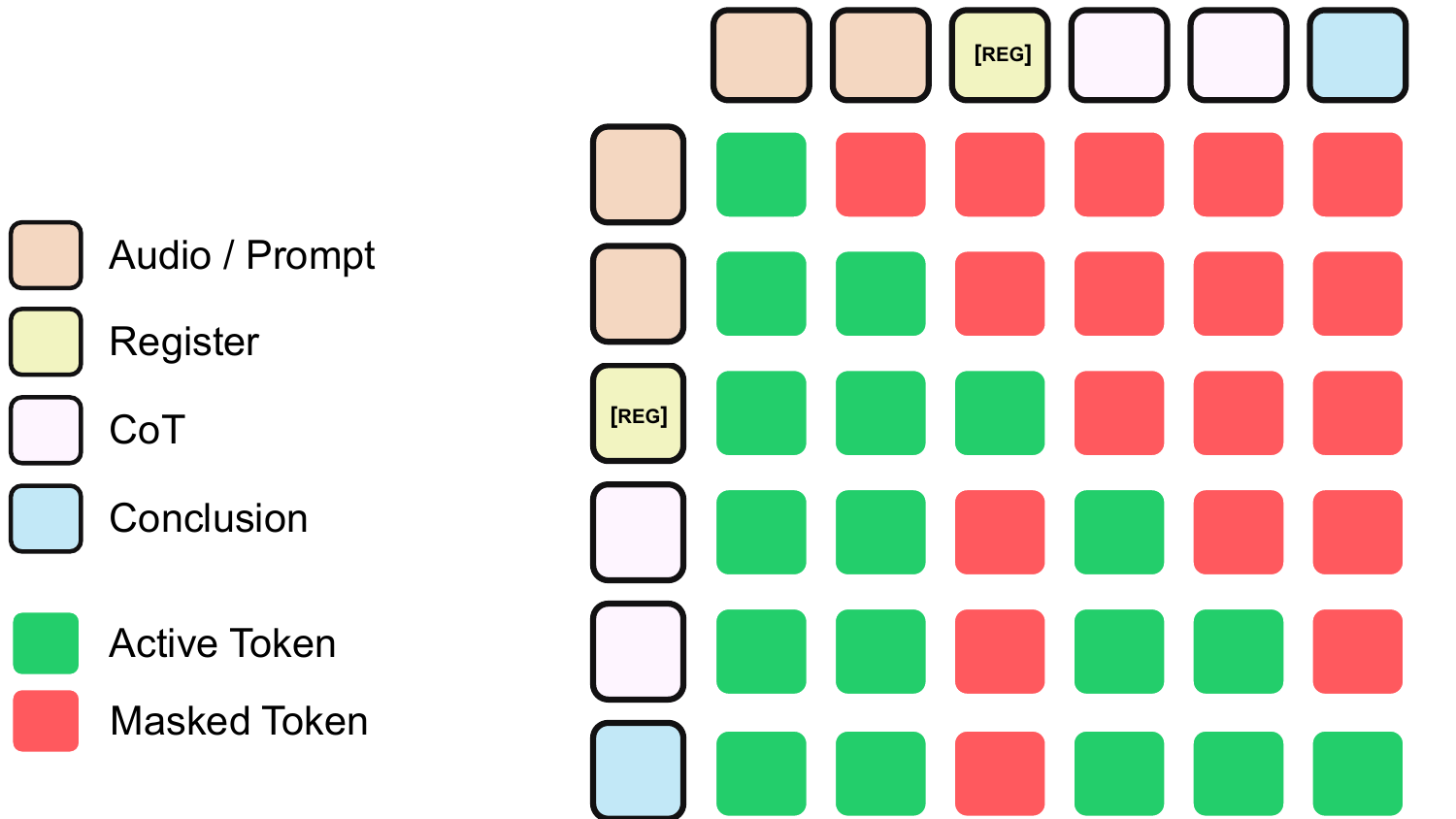}
    \caption{Custom Causal Attention Masking. The [REG] token attends to preceding context, but subsequent tokens cannot attend to it. This architecture preserves standard, modification-free SFT inference behavior.}
    \label{fig:attention_mask}
\end{figure}

Notably, the register token does not influence the context of other tokens. This ensures our approach maintains the same inference dynamics as standard SFT, as illustrated in Fig. \ref{fig:attention_mask}. Fig. ~\ref{fig:methodology_overview} provides a schematic overview of the proposed SPARE framework.

\subsection{SPARE: Latent Conclusion Alignment}

The core of the SPARE method is the alignment of the register's hidden state with the terminal conclusion. We extract the text within the Conclusion tags and pass it through a frozen \textbf{Sentence BERT} encoder \footnote{https://huggingface.co/sentence-transformers/all-MiniLM-L6-v2} to obtain a target semantic embedding $v_{target}$.

Let $h_{[REG]}$ be the last hidden state of the register token at the final layer of the LLM. We define the alignment loss $\mathcal{L}_{align}$ using cosine distance:

\[
\mathcal{L}_{align} = 1 - \frac{h_{[REG]} \cdot v_{target}}{\parallel h_{[REG]} \parallel \parallel v_{target}\parallel}
\]

The total training objective is a weighted sum of the standard cross-entropy loss $\mathcal{L}_{CE}$ and our alignment objective:

$$\mathcal{L}_{total} = \mathcal{L}_{CE} + \lambda \cdot \mathcal{L}_{align}$$

where $\lambda = 2.0$ was determined to be the optimal balance through empirical validation.

\subsection{Inference and Model Properties}


Beyond improving audio reasoning, a key advantage of SPARE is its efficiency and structural compatibility. Rather than functioning as a standalone alternative architecture, SPARE is a complementary fine-tuning strategy that layer-caches seamlessly onto existing backbones. At inference time, the model follows standard autoregressive decoding, and both the $[REG]$ token and the projection head are completely discarded. Consequently, the model requires no additional parameters, introducing zero computational overhead or latency penalties during deployment. The performance gains observed in zero-shot evaluations are therefore attributed solely to the internal regularization provided by the alignment loss during training, which encourages the model to ground its initial latent states in the ultimate semantic goal.

\section{Experimental Setup}

\subsection{Comparison with Baselines}

We evaluate \textbf{SPARE} against four primary configurations using  SALMONN 13B as a base model: the Zero-shot, Zero-shot (CoT), a standard Supervised Fine-Tuning (SFT) baseline on the AF-Think dataset, and an Audio MuToR implementation, inspired from \cite{gerontopoulos2025}. MuToR represents a multi-token prediction approach adapted for audio, that makes use of register token, injecting future knowledge into the models, serving as a competitive baseline for structured intermediate representations.

We utilize two primary benchmarks that challenge the model beyond simple sound recognition:

\begin{itemize}
    \item \textbf{MMAU (Multi-Modal Audio Understanding)} \cite{sakshi2024mmaumassivemultitaskaudio}: This benchmark focuses on complex audio tasks that require high-level comprehension and multi-step deduction. It assesses the model's ability to integrate diverse acoustic cues, such as speech, environmental sounds, and musical elements, to answer questions that demand a holistic understanding of the audio scene.
    \item \textbf{MMAR (Multi-Modal Audio Reasoning)} \cite{ma2025mmar}: Specifically designed to test logical consistency and the "Chain-of-Thought" capabilities of audio-language models, MMAR provides a rigorous testbed for logical inference. It evaluates whether a model can maintain a coherent reasoning path from the initial acoustic input to the final conclusion without succumbing to "audio drift" or hallucinatory reasoning.

\end{itemize}

\section{Results and Discussion}

As shown in Table \ref{tab:main_results}, SPARE achieves the highest performance across both benchmarks, reaching \textbf{58.03\% on MMAU} and \textbf{40.32\% on MMAR}. Notably, our method provides a significant boost over SFT (+3.38\% on MMAU) and comfortably outperforms Audio MuToR (+5.01\% on MMAU). This suggests that the latent conclusion alignment effectively ``seeds" the model with a global goal, preventing the logical drift common in standard SFT and even in more complex multi-token prediction schemes like MuToR.

While high-resource models like Audio Flamingo 3 (73.30\% on MMAU) and Qwen2.5-Omni (71.50\%) achieve higher scores, they rely on massive datasets and industrial-scale compute. Our goal is not to compete with these through scaling, but to propose a novel, efficient methodology. By significantly outperforming strong baselines on a controlled dataset, we demonstrate that SPARE enhances reasoning and auditory grounding without the need for massive-scale training.

\subsection{Ablation Studies and Sensitivity Analysis}

We conducted extensive ablations to investigate the impact of the alignment weight $\lambda$ and the architectural placement of the register tokens.

\textbf{Impact of Alignment Weight:} Table \ref{tab:ablations} demonstrates that SPARE is relatively robust to the choice of $\lambda$. While performance peaks at $\lambda=2.0$, values of 1.0 and 3.0 still yield results superior to all non-SPARE baselines. However, at $\lambda=3.0$, we observe a higher variance ($\pm 2.83$ on MMAU), suggesting that overly aggressive regularization may introduce instability in the latent space during training.


\textbf{Register Token Placement:} We evaluated injecting register tokens before every sequence chapter (``Multi-conclusion summary'') and aligning each with its respective Sentence-BERT embedding (``Chapters summary''). Both variants degraded performance, dropping to 55.43\% on MMAU, and introduced severe training instability. This confirms that a single sequence-initial bottleneck is superior, forcing the model to extract a unified, grounded acoustic representation before decoding begins.

\subsection{Comparison with Audio MuToR}

The results for \textbf{Audio MuToR} (Table \ref{tab:ablations}) show that while increasing the prediction distance ($d_{max}$) and applying $\lambda$ decay can improve stability, the method still fails to match the reasoning accuracy of SPARE. We attribute this to the fact that MuToR focuses on predicting intermediate segments (syntactic nuances), whereas SPARE aligns the model directly with the terminal semantic conclusion. By supervising the start of the reasoning chain with the ultimate goal, SPARE ensures a higher degree of global consistency that segment-level prediction cannot achieve.

\begin{table}[t]
\caption{Zero-shot accuracy (\%) on MMAR and MMAU. All evaluations utilize SALMONN 13B as the backbone. Performance is compared across five configurations: (1) Zero-shot: immediate conclusion; (2) Zero-shot (CoT): Chain-of-Thought reasoning; (3) SFT: supervised fine-tuning on YouTube8m; (4) Audio MuToR: MuToR adaptation fine-tuned on YouTube8m; and (5) SPARE: our proposed approach fine-tuned on YouTube8m. SPARE outperforms all baselines by a significant margin.}
\label{tab:main_results}
\centering
\begin{tabular}{@{}lcc@{}}
\toprule
\textbf{Method} & \textbf{MMAU (\%)} & \textbf{MMAR (\%)} \\ \midrule
Zero-shot & 36.02 ± 1.02 & 33.28 ± 0.75 \\
Zero-shot (CoT) & 18.03 ± 0.36 & 12.32 ± 2.31 \\
SFT & 54.65 ± 0.34 & 38.15 ± 0.37 \\
Audio MuToR & 53.02 ± 1.23 & 38.40 ± 0.53 \\
\textbf{SPARE (Ours)} & \textbf{58.03 ± 1.50} & \textbf{40.32 ± 0.57} \\ \bottomrule
\end{tabular}
\end{table}

\begin{table}[t]
\caption{Ablation studies for SPARE and Audio MuToR (Zero-shot \% accuracy). Results on MMAR and MMAU datasets. For SPARE, we ablate $\lambda$; for Audio MuToR, we vary $\lambda$, $d_{min}$, and $d_{max}$. The ``Multi-conclusion summary" and ``Chapter summary" variants utilize register tokens before every chapter rather than just sequence-initially. (*) denotes the application of $\lambda$ decay.}
\label{tab:ablations}
\centering
\begin{tabular}{@{}lcc@{}}
\toprule
\textbf{Method} & \textbf{MMAU (\%)} & \textbf{MMAR (\%)} \\ \midrule
SPARE (Ours) & & \\
\midrule
$\lambda$ = 1.0 & 57.73 ± 1.77 & 39.72 ± 1.09 \\
\textbf{$\lambda$ = 2.0} & \textbf{58.03 ± 1.50} & \textbf{40.32 ± 0.57} \\ 
$\lambda$ = 3.0 & 57.60 ± 2.83 & 39.92 ± 1.04 \\ 
Multi-conclusion summary & 55.43 ± 0.12 & 38.60 ± 0.25 \\ 
Chapters summary & 52.18 ± 8.95 & 38.62 ± 2.18 \\ \midrule \midrule
Audio MuToR & & \\
\midrule
$\lambda$ = 0.1, $d_{min}$ = 2, $d_{max}$ = 4 & 52.87 ± 0.56 & 38.25 ± 1.56 \\
\textbf{$\lambda$ = 0.1, $d_{min}$ = 2, $d_{max}$ = 20 *} & \textbf{53.02 ± 1.23} & \textbf{38.40 ± 0.53} \\
$\lambda$ = 0.05, $d_{min}$ = 2, $d_{max}$ = 30 * & 53.58 ± 0.16 & 37.24 ± 15.03 \\
\bottomrule
\end{tabular}
\end{table}

\begin{figure*}[t]
  \centering
  \vspace{-20pt}
  \begin{subfigure}[t]{0.4\linewidth} 
    \centering
    \includegraphics[width=\linewidth]{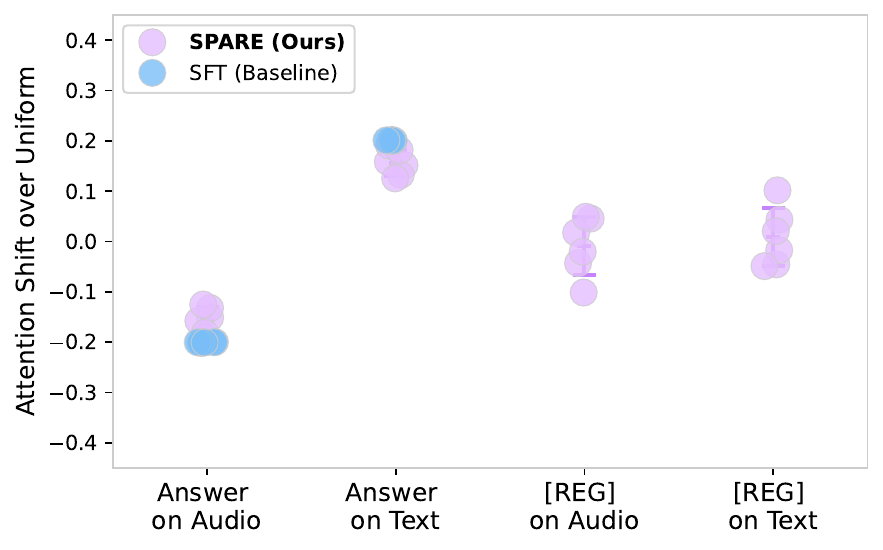}
    \subcaption{First Layer}
    \label{subfig:attention_maps_a}
  \end{subfigure}
  \hspace{2em} 
  \begin{subfigure}[t]{0.4\linewidth} 
    \centering
    \includegraphics[width=\linewidth]{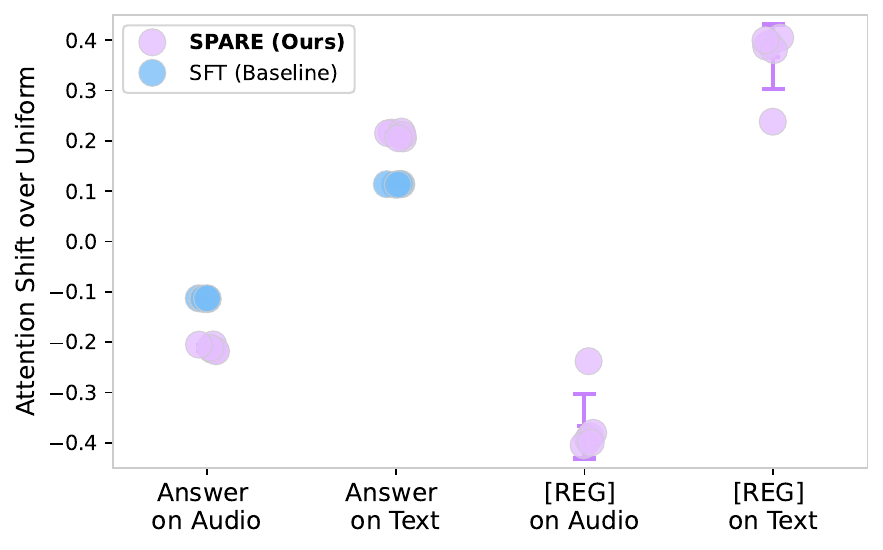}
    \subcaption{Last Layer}
    \label{subfig:attention_maps_b}
  \end{subfigure}
  \caption{We computed an attention score for SPARE (pink) and the SFT baseline (blue) across the First (a) and Last (b) layers. The y-axis measures the attention shift over a uniform baseline ($\frac{1}{L}$), where values above 0.0 indicate a focus stronger than average and values below indicate a relative lack of focus. The x-axis evaluates attention for both the generated answer and our injected [REG] token, specifically assessing the average attention of all generated answer tokens toward audio frames or text tokens (question + choices + answer), alongside the specific attention of the register token toward those same targets.}
  \label{fig:attention_maps}
\end{figure*}


\subsection{Analysis of Attention Maps}

A persistent challenge in LALMs is the tendency of the self-attention mechanism to prioritize linguistic context over acoustic features, often leading to a ``reasoning gap" where the model ignores the source audio during long-form generation \cite{ahia2025blabbrutallylongaudio, yang_audiolens_2025, wang_pay_2025}. We observed this phenomenon in our baseline configurations, where attention was primarily dominated by preceding text tokens rather than the input audio signal. To investigate how SPARE mitigates this, we performed a mechanistic analysis of the self-attention maps across six independent random seeds.

We analyzed the self-attention weights of both the generated answer and the register tokens ([REG]) relative to the audio segments across the first and last layers of the LLM backbone, as illustrated in Fig. \ref{fig:attention_maps}. Our results reveal that in the initial layer of the SPARE model, [REG] tokens exhibit significantly higher attention scores toward audio frames than the autoregressively generated answer tokens. This suggests that to align with the terminal semantic goal (the Sentence BERT conclusion embedding) the register token must prioritize the extraction of meaningful acoustic content.
This shift appears to encourage a localized increase in auditory focus across the entire sequence; notably, answer tokens in the SPARE model demonstrate enhanced attention toward the audio compared to the SFT baseline. This mechanism facilitates the extraction of relevant audio features at the sequence's start, which are subsequently integrated into the linguistic domain by the final layer. Such a transition from the acoustic to the textual domain aligns with the model's core objective of grounded reasoning~\cite{mousavi2025alas,yang2025training}.
These results provide empirical evidence that our approach acts as a structural regularizer, improving auditory grounding by ensuring the model ``listens" more intently to the audio before committing to a reasoning path.

Our mechanistic findings are further corroborated by qualitative evaluations\footnote{https://my-demo-hub.github.io/spare/}. When prompted to listen closely to the acoustic input, SPARE consistently generates longer, more detailed, and tightly grounded audio captions compared to baseline configurations.

\section{Conclusions}


We introduced SPARE, a latent regularization strategy designed to bridge the reasoning gap in Large Audio Language Models. By aligning a register token with the semantic embedding of the final conclusion, SPARE encourages the model to form a grounded, goal-oriented representation of the audio before decoding begins. Experiments on the MMAU and MMAR benchmarks show that SPARE significantly reduces the performance gap between direct-answer prompting and CoT reasoning compared to standard SFT and Audio MuToR. Attention analysis further reveals that the alignment objective encourages stronger focus on acoustic features in early layers, helping mitigate audio drift during long reasoning chains. These results demonstrate that latent summary alignment can effectively improve audio-grounded reasoning without additional inference cost.

\section{Acknowledgments}
We sincerely thank Jihoon Jeong for assistance with dataset curation, Pietro Cagnasso for creating Figure~\ref{fig:attention_maps}, Gianfranco Dumoulin Bertucci for reviewing the code, and Francesco Paissan for insightful discussions. This work was supported by research funding, computational resources, and donations from NSERC, the Digital Research Alliance of Canada (alliancecan.ca), the Translated Imminent Program, and an Apple Seed Grant.

\section{Generative AI Use Disclosure}
During the preparation of this work, the authors utilized generative AI tools for the purposes of manuscript editing and linguistic polishing. The AI was not used to produce a significant part of the manuscript’s core technical contributions, data analysis, or original conclusions. All (co-)authors have reviewed and edited the generated output, remain fully responsible and accountable for the content of the paper, and consent to its submission in accordance with ISCA policy.

\bibliographystyle{IEEEtran}
\bibliography{mybib}

@misc{gerontopoulos2025,
      title={Multi-Token Prediction Needs Registers}, 
      author={Anastasios Gerontopoulos and Spyros Gidaris and Nikos Komodakis},
      year={2025},
      eprint={2505.10518},
      archivePrefix={arXiv},
      primaryClass={cs.CL},
      url={https://arxiv.org/abs/2505.10518}, 
}

@misc{mahajan2025multitokenpredictionpretrainingllms,
      title={Beyond Multi-Token Prediction: Pretraining LLMs with Future Summaries}, 
      author={Divyat Mahajan and Sachin Goyal and Badr Youbi Idrissi and Mohammad Pezeshki and Ioannis Mitliagkas and David Lopez-Paz and Kartik Ahuja},
      year={2025},
      eprint={2510.14751},
      archivePrefix={arXiv},
      primaryClass={cs.LG},
      url={https://arxiv.org/abs/2510.14751}, 
}

@misc{tang_salmonn_2024,
	title = {{SALMONN}: {Towards} {Generic} {Hearing} {Abilities} for {Large} {Language} {Models}},
	shorttitle = {{SALMONN}},
	url = {http://arxiv.org/abs/2310.13289},
	doi = {10.48550/arXiv.2310.13289},
	urldate = {2025-12-22},
	publisher = {arXiv},
	author = {Tang, Changli and Yu, Wenyi and Sun, Guangzhi and Chen, Xianzhao and Tan, Tian and Li, Wei and Lu, Lu and Ma, Zejun and Zhang, Chao},
	month = apr,
	year = {2024},
	note = {arXiv:2310.13289 [cs]},
}

@inproceedings{
gogoel_audio_2025,
title={Audio Flamingo 3: Advancing Audio Intelligence with Fully Open Large Audio Language Models},
author={Sreyan Ghosh and Arushi Goel and Jaehyeon Kim and Sonal Kumar and Zhifeng Kong and Sang-gil Lee and Chao-Han Huck Yang and Ramani Duraiswami and Dinesh Manocha and Rafael Valle and Bryan Catanzaro},
booktitle={The Thirty-ninth Annual Conference on Neural Information Processing Systems},
year={2025},
url={https://openreview.net/forum?id=FjByDpDVIO}
}

@misc{xu_qwen25-omni_2025,
	title = {Qwen2.5-{Omni} {Technical} {Report}},
	url = {http://arxiv.org/abs/2503.20215},
	doi = {10.48550/arXiv.2503.20215},
	urldate = {2026-02-26},
	publisher = {arXiv},
	author = {Xu, Jin and Guo, Zhifang and He, Jinzheng and Hu, Hangrui and He, Ting and Bai, Shuai and Chen, Keqin and Wang, Jialin and Fan, Yang and Dang, Kai and Zhang, Bin and Wang, Xiong and Chu, Yunfei and Lin, Junyang},
	month = mar,
	year = {2025},
	note = {arXiv:2503.20215 [cs]}
}

@misc{xie_audio-reasoner_2025,
	title = {Audio-{Reasoner}: {Improving} {Reasoning} {Capability} in {Large} {Audio} {Language} {Models}},
	shorttitle = {Audio-{Reasoner}},
	url = {http://arxiv.org/abs/2503.02318},
	doi = {10.48550/arXiv.2503.02318},
	urldate = {2026-01-08},
	publisher = {arXiv},
	author = {Xie, Zhifei and Lin, Mingbao and Liu, Zihang and Wu, Pengcheng and Yan, Shuicheng and Miao, Chunyan},
	month = sep,
	year = {2025},
	note = {arXiv:2503.02318 [cs]}
}

@misc{wu_audio-thinker_2025,
	title = {Audio-{Thinker}: {Guiding} {Audio} {Language} {Model} {When} and {How} to {Think} via {Reinforcement} {Learning}},
	shorttitle = {Audio-{Thinker}},
	url = {http://arxiv.org/abs/2508.08039},
	doi = {10.48550/arXiv.2508.08039},
	urldate = {2026-02-26},
	publisher = {arXiv},
	author = {Wu, Shu and Li, Chenxing and Wang, Wenfu and Zhang, Hao and Wang, Hualei and Yu, Meng and Yu, Dong},
	month = aug,
	year = {2025},
	note = {arXiv:2508.08039 [cs]
version: 1},
}

@misc{radford_robust_2022,
	title = {Robust {Speech} {Recognition} via {Large}-{Scale} {Weak} {Supervision}},
	url = {http://arxiv.org/abs/2212.04356},
	doi = {10.48550/arXiv.2212.04356},
	urldate = {2026-02-25},
	publisher = {arXiv},
	author = {Radford, Alec and Kim, Jong Wook and Xu, Tao and Brockman, Greg and McLeavey, Christine and Sutskever, Ilya},
	month = dec,
	year = {2022},
	note = {arXiv:2212.04356 [eess]},
}

@misc{chen_beats_2022-1,
	title = {{BEATs}: {Audio} {Pre}-{Training} with {Acoustic} {Tokenizers}},
	shorttitle = {{BEATs}},
	url = {http://arxiv.org/abs/2212.09058},
	doi = {10.48550/arXiv.2212.09058},
	urldate = {2026-02-25},
	publisher = {arXiv},
	author = {Chen, Sanyuan and Wu, Yu and Wang, Chengyi and Liu, Shujie and Tompkins, Daniel and Chen, Zhuo and Wei, Furu},
	month = dec,
	year = {2022},
	note = {arXiv:2212.09058 [eess]},
}

@misc{yang_audiolens_2025,
	title = {{AudioLens}: {A} {Closer} {Look} at {Auditory} {Attribute} {Perception} of {Large} {Audio}-{Language} {Models}},
	shorttitle = {{AudioLens}},
	url = {http://arxiv.org/abs/2506.05140},
	doi = {10.48550/arXiv.2506.05140},
	urldate = {2026-03-02},
	publisher = {arXiv},
	author = {Yang, Chih-Kai and Ho, Neo and Lee, Yi-Jyun and Lee, Hung-yi},
	month = aug,
	year = {2025},
	note = {arXiv:2506.05140 [cs]},
}

@misc{wang_pay_2025,
	title = {Pay {More} {Attention} {To} {Audio}: {Mitigating} {Imbalance} of {Cross}-{Modal} {Attention} in {Large} {Audio} {Language} {Models}},
	shorttitle = {Pay {More} {Attention} {To} {Audio}},
	url = {http://arxiv.org/abs/2509.18816},
	doi = {10.48550/arXiv.2509.18816},
	urldate = {2026-03-02},
	publisher = {arXiv},
	author = {Wang, Junyu and Ma, Ziyang and Luo, Zhengding and Wang, Tianrui and Ge, Meng and Wang, Xiaobao and Wang, Longbiao},
	month = sep,
	year = {2025},
	note = {arXiv:2509.18816 [cs]},
}

@misc{sakshi2024mmaumassivemultitaskaudio,
      title={MMAU: A Massive Multi-Task Audio Understanding and Reasoning Benchmark}, 
      author={S Sakshi and Utkarsh Tyagi and Sonal Kumar and Ashish Seth and Ramaneswaran Selvakumar and Oriol Nieto and Ramani Duraiswami and Sreyan Ghosh and Dinesh Manocha},
      year={2024},
      eprint={2410.19168},
      archivePrefix={arXiv},
      primaryClass={eess.AS},
      url={https://arxiv.org/abs/2410.19168}, 
}

@article{ma2025mmar,
  title={MMAR: A Challenging Benchmark for Deep Reasoning in Speech, Audio, Music, and Their Mix},
  author={Ma, Ziyang and Ma, Yinghao and Zhu, Yanqiao and Yang, Chen and Chao, Yi-Wen and Xu, Ruiyang and others},
  journal={Proc. NeurIPS},
  year={2025}
}

@misc{wang2020minilmdeepselfattentiondistillation,
      title={MiniLM: Deep Self-Attention Distillation for Task-Agnostic Compression of Pre-Trained Transformers}, 
      author={Wenhui Wang and Furu Wei and Li Dong and Hangbo Bao and Nan Yang and Ming Zhou},
      year={2020},
      eprint={2002.10957},
      archivePrefix={arXiv},
      primaryClass={cs.CL},
      url={https://arxiv.org/abs/2002.10957}, 
}

@misc{ahia2025blabbrutallylongaudio,
      title={BLAB: Brutally Long Audio Bench},
      author={Orevaoghene Ahia and Martijn Bartelds and Kabir Ahuja and Hila Gonen and Valentin Hofmann and Siddhant Arora and Shuyue Stella Li and Vishal Puttagunta and Mofetoluwa Adeyemi and Charishma Buchireddy and Ben Walls and Noah Bennett and Shinji Watanabe and Noah A. Smith and Yulia Tsvetkov and Sachin Kumar},
      year={2025},
      eprint={2505.03054},
      archivePrefix={arXiv},
      primaryClass={cs.AI},
      url={https://arxiv.org/abs/2505.03054},
}

@article{comanici2025gemini,
  title={Gemini 2.5: Pushing the frontier with advanced reasoning, multimodality, long context, and next generation agentic capabilities},
  author={Comanici, Gheorghe and Bieber, Eric and Schaekermann, Mike and Pasupat, Ice and Sachdeva, Noveen and Dhillon, Inderjit and Blistein, Marcel and Ram, Ori and Zhang, Dan and Rosen, Evan and others},
  journal={arXiv preprint arXiv:2507.06261},
  year={2025}
}

@inproceedings{zhouleast,
  title={Least-to-Most Prompting Enables Complex Reasoning in Large Language Models},
  author={Zhou, Denny and Sch{\"a}rli, Nathanael and Hou, Le and Wei, Jason and Scales, Nathan and Wang, Xuezhi and Schuurmans, Dale and Cui, Claire and Bousquet, Olivier and Le, Quoc V and others},
  booktitle={The Eleventh International Conference on Learning Representations}
}

@misc{darcet2024visiontransformersneedregisters,
      title={Vision Transformers Need Registers}, 
      author={Timothée Darcet and Maxime Oquab and Julien Mairal and Piotr Bojanowski},
      year={2024},
      eprint={2309.16588},
      archivePrefix={arXiv},
      primaryClass={cs.CV},
      url={https://arxiv.org/abs/2309.16588}, 
}

@article{yao2023tree,
  title={Tree of thoughts: Deliberate problem solving with large language models},
  author={Yao, Shunyu and Yu, Dian and Zhao, Jeffrey and Shafran, Izhak and Griffiths, Tom and Cao, Yuan and Narasimhan, Karthik},
  journal={Advances in neural information processing systems},
  volume={36},
  pages={11809--11822},
  year={2023}
}

@article{wu2025step,
  title={Step-audio 2 technical report},
  author={Wu, Boyong and Yan, Chao and Hu, Chen and Yi, Cheng and Feng, Chengli and Tian, Fei and Shen, Feiyu and Yu, Gang and Zhang, Haoyang and Li, Jingbei and others},
  journal={arXiv preprint arXiv:2507.16632},
  year={2025}
}

@article{li2025reinforcement,
  title={Reinforcement learning outperforms supervised fine-tuning: A case study on audio question answering},
  author={Li, Gang and Liu, Jizhong and Dinkel, Heinrich and Niu, Yadong and Zhang, Junbo and Luan, Jian},
  journal={arXiv preprint arXiv:2503.11197},
  year={2025}
}

@inproceedings{yang2025sakura,
  title={SAKURA: On the Multi-hop Reasoning of Large Audio-Language Models Based on Speech and Audio Information},
  author={Yang, Chih-Kai and Ho, Neo and Piao, Yen-Ting and Lee, Hung-yi},
  booktitle={Proc. Interspeech 2025},
  pages={1788--1792},
  year={2025}
}

@article{chen2025audio,
  title={Do Audio LLMs Really LISTEN, or Just Transcribe? Measuring Lexical vs. Acoustic Emotion Cues Reliance},
  author={Chen, Jingyi and Guo, Zhimeng and Chun, Jiyun and Wang, Pichao and Perrault, Andrew and Elsner, Micha},
  journal={arXiv preprint arXiv:2510.10444},
  year={2025}
}

@article{xu2025qwen3,
  title={Qwen3-omni technical report},
  author={Xu, Jin and Guo, Zhifang and Hu, Hangrui and Chu, Yunfei and Wang, Xiong and He, Jinzheng and Wang, Yuxuan and Shi, Xian and He, Ting and Zhu, Xinfa and others},
  journal={arXiv preprint arXiv:2509.17765},
  year={2025}
}

@article{li2025baichuan,
  title={Baichuan-omni-1.5 technical report},
  author={Li, Yadong and Liu, Jun and Zhang, Tao and Chen, Song and Li, Tianpeng and Li, Zehuan and Liu, Lijun and Ming, Lingfeng and Dong, Guosheng and Pan, Da and others},
  journal={arXiv preprint arXiv:2501.15368},
  year={2025}
}

@article{deshmukh2025mellow,
  title={Mellow: a small audio language model for reasoning},
  author={Deshmukh, Soham and Dixit, Satvik and Singh, Rita and Raj, Bhiksha},
  journal={arXiv preprint arXiv:2503.08540},
  year={2025}
}

@article{ma2025audio,
  title={Audio-cot: Exploring chain-of-thought reasoning in large audio language model},
  author={Ma, Ziyang and Chen, Zhuo and Wang, Yuping and Chng, Eng Siong and Chen, Xie},
  journal={arXiv preprint arXiv:2501.07246},
  year={2025}
}

@article{wei2022chain,
  title={Chain-of-thought prompting elicits reasoning in large language models},
  author={Wei, Jason and Wang, Xuezhi and Schuurmans, Dale and Bosma, Maarten and Xia, Fei and Chi, Ed and Le, Quoc V and Zhou, Denny and others},
  journal={Advances in neural information processing systems},
  volume={35},
  pages={24824--24837},
  year={2022}
}

@article{kojima2022large,
  title={Large language models are zero-shot reasoners},
  author={Kojima, Takeshi and Gu, Shixiang Shane and Reid, Machel and Matsuo, Yutaka and Iwasawa, Yusuke},
  journal={Advances in neural information processing systems},
  volume={35},
  pages={22199--22213},
  year={2022}
}

@article{mousavi2025alas,
  title={ALAS: Measuring Latent Speech-Text Alignment For Spoken Language Understanding In Multimodal LLMs},
  author={Mousavi, Pooneh and Wang, Yingzhi and Ravanelli, Mirco and Subakan, Cem},
  journal={arXiv preprint arXiv:2505.19937},
  year={2025}
}

@inproceedings{yang2025training,
  title={Training Strategies for Speech Large Language Models: A Comprehensive Survey},
  author={Yang, Shiqi and Huang, Ziyi and Xiao, Wengran},
  booktitle={2025 3rd International Conference on Foundation and Large Language Models (FLLM)},
  pages={161--171},
  year={2025},
  organization={IEEE}
}

\end{document}